%% file: acl_latex.tex
\documentclass[11pt]{article}

\usepackage[preprint]{acl}
\usepackage[table]{xcolor}
\usepackage{graphicx} 
\usepackage{times}
\usepackage{latexsym}

\usepackage[font=small,skip=3pt]{caption}
\usepackage{booktabs}
\usepackage{algorithm}
\usepackage{booktabs}
\usepackage{multirow}
\usepackage{amsmath}
\usepackage{algpseudocode}
\algrenewcommand\algorithmicrequire{\textbf{Input:}}
\algrenewcommand\algorithmicensure{\textbf{Output:}}
\algtext*{EndIf} 
\algtext*{EndFor} 
\algtext*{EndWhile}
\algtext*{EndProcedure}
\usepackage{pdfpages}
\usepackage{graphicx}
\usepackage{caption}
\usepackage{subcaption}

\usepackage[T1]{fontenc}

\usepackage[utf8]{inputenc}

\usepackage{microtype}

\usepackage{inconsolata}

\usepackage[most]{tcolorbox}
\tcbuselibrary{breakable} 

\usepackage{amssymb}
\usepackage{xcolor}
\usepackage{bm}
\usepackage{makecell}
\usepackage{enumitem}
\setlist[itemize]{nosep, topsep=2pt}
\setlist[itemize]{leftmargin=*}
\usepackage{algorithm}  
\usepackage{algorithmicx}  
\usepackage{amsthm}

\newtcolorbox{promptdisplaybox}[1][]{
    colback=gray!10,       
    colframe=black,        
    fontupper={\small},    
    breakable,             
    enhanced,              
    left=2mm,              
    right=2mm,             
    top=1mm,               
    bottom=1mm,            
    #1                     
}

\title{Aggregation Queries over Unstructured Text:\\Benchmark and Agentic Method}

\makeatletter

\newcommand{\Rmnum}[1]{\expandafter\@slowromancap\romannumeral #1@}
\makeatother

\author{
Haojia Zhu$^1$,
Qinyuan Xu$^2$,
Haoyu Li$^1$,
Yuxi Liu$^1$,
Hanchen Qiu$^1$,
Jiaoyan Chen$^3$,
Jiahui Jin$^1$\\
$^1$Southeast University, $^2$Imperial College London, $^3$The University of Manchester\\
zhuhaojia@seu.edu.cn, qx225@ic.ac.uk, \{213233043, 213231903, 220246359\}@seu.edu.cn,\\ jiaoyan.chen.manchester.ac.uk, jjin@seu.edu.cn
}

\begin{document}
\maketitle
\begin{abstract}
Aggregation query over free text is a long-standing yet underexplored problem. Unlike ordinary question answering, aggregate queries require exhaustive evidence collection and systems are required to “find all,” not merely “find one.” Existing paradigms such as Text-to-SQL and Retrieval-Augmented Generation fail to achieve this completeness. In this work, we formalize entity-level aggregation querying over text in a corpus-bounded setting with strict completeness requirement. To enable principled evaluation, we introduce \textsc{AGGBench}, a benchmark designed to evaluate completeness-oriented aggregation under realistic large-scale corpus. To accompany the benchmark, we propose \textsc{DFA} (Disambiguation--Filtering--Aggregation), a modular agentic baseline that decomposes aggregation querying into interpretable stages and exposes key failure modes related to ambiguity, filtering, and aggregation. Empirical results show that DFA consistently improves aggregation evidence coverage over strong RAG and agentic baselines. The data and code are available in \url{https://anonymous.4open.science/r/DFA-A4C1}.
\end{abstract}

\input{sections/Introduction}

\input{sections/Related_Work}

\input{sections/Preliminaries}

\input{sections/Benchmark}
\input{sections/Methodology}

\input{sections/Experiments}

\input{sections/Conclusion}
\clearpage
\input{sections/limitations}
\input{sections/ethical_impact}


\bibliography{custom}

\appendix
\input{sections/appendix}

\end{document}

%% file: sections/Introduction.tex
\section{Introduction} \label{sec:intro}

\begin{figure}[t]
  \centering
  \includegraphics[width=0.92\linewidth, trim=0 0 0 0, clip]{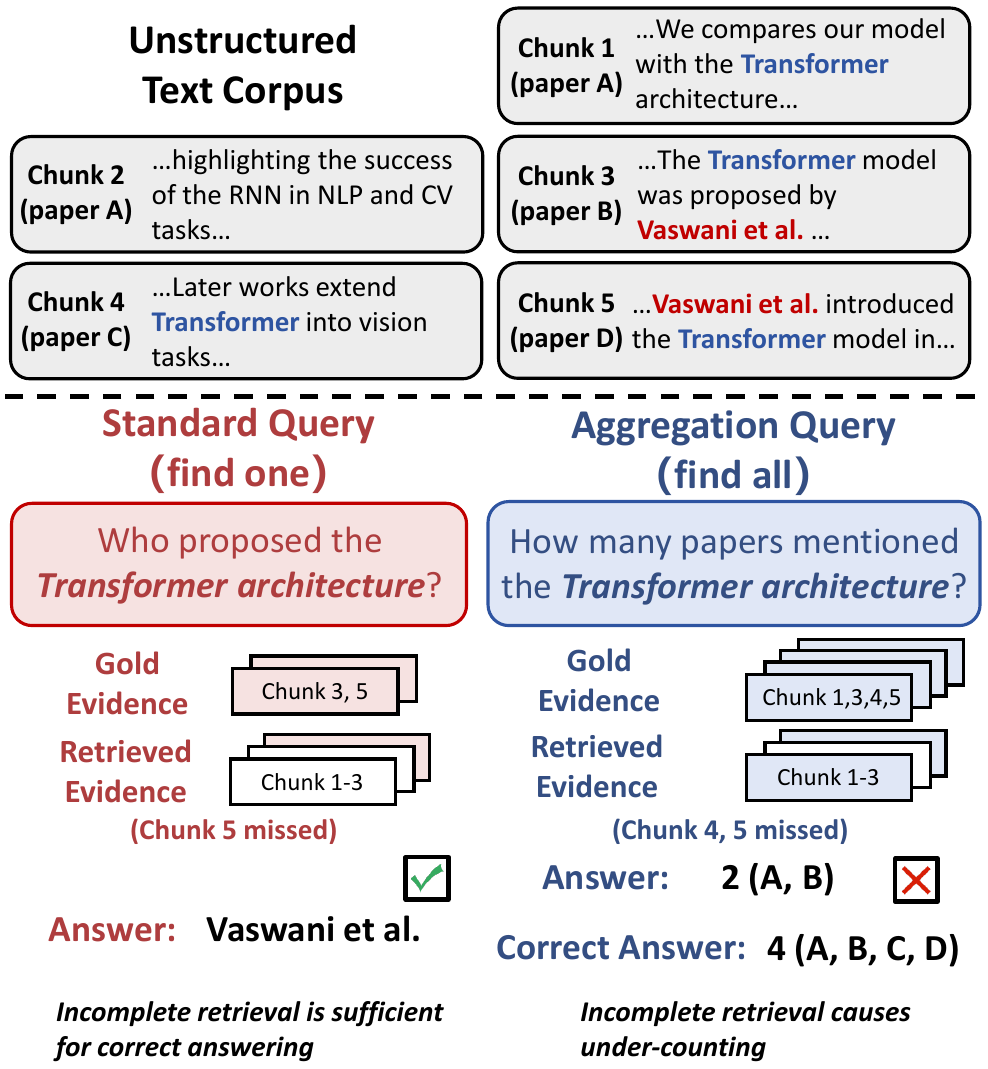}
  \caption{Examples of Aggregation Query and Standard Query over Unstructured Data.}
  \label{fig:intro}
  \vspace{-4mm}
\end{figure}

Aggregation queries arise in many real-world analytical tasks that require systematic inspection of large text collections, e.g., counting how many entities in a corpus satisfy a set of regulatory or factual conditions. Such needs are common in domains including law~\cite{niklaus-etal-2024-multilegalpile}, finance~\cite{xie-etal-2024-finben}, compliance~\cite{jain-etal-2025-regulatory-compliance}, and investigative analysis~\cite{spangher-etal-2025-journalist-source-selection}. For example, legal e-discovery and contract analytics must locate all clauses that match specific regulatory or contractual patterns~\cite{wang-etal-2025-acord}. Financial and compliance audit require enumerating entities exhibiting risk signals in thousands of files~\cite{li-etal-2025-fingear}. Recent data-analysis agents~\cite{yao2023react,shen2023hugginggpt,zhou2024languageagenttreesearch} also increasingly aim to analyze large textual corpora, where filtering and entity grounding form the backbone of many tasks. As shown in Figure~\ref{fig:intro}, unlike standard QA tasks that aim to “find” a plausible answer, aggregation querying is more challenging because it requires "find all" entities that satisfy the specified conditions.

Aggregation querying over text has not been explicitly studied as a standalone problem, but there are some existing solutions that are applicable to this problem: (1) Structure and schema-first transformation. The most common approach converts the text corpus into relational or graph databases and applies Text-to-SQL or symbolic queries~\cite{yu-etal-2018-spider, zhong2017seq2sql}. However, recent work in both academic and industrial practice has found that constructing such structured representations requires costly preprocessing pipelines, and the resulting structured repositories are often of low quality~\cite{katti2021weblists, siino-2024-preprocessing}. It also enforces a fixed schema that limits flexibility for schema-free or compositional natural-language queries. As a result, these approaches are rarely feasible for large, heterogeneous, or continuously evolving corpora. (2) Retrieval-augmented reasoning. RAG frameworks~\cite{lewis-etal-2020-rag, asai2023selfrag} offer an alternative by letting LLMs reason directly over the retrieved passages. But RAG was designed for open-domain QA, where the goal is to retrieve some highly relevant evidence, not to obtain complete coverage. Their fixed top-k retrieval window~\cite{karpukhin-etal-2020-dense} makes them fundamentally incompatible with aggregation. The relevant mentions often exceed k, and omissions in retrieval propagate irreparably into the reasoning stage, leading to systematic undercounting or incomplete aggregation. 

This gap at the method level is further compounded by a corresponding gap in evaluation: existing benchmarks are not designed to isolate or diagnose aggregation-specific failure modes. Existing QA benchmarks focus on factoid or open-domain retrieval tasks rather than aggregation reasoning. Datasets such as HotpotQA~\cite{yang-etal-2018-hotpotqa} and WikiMultiHopQA~\cite{ho-etal-2020-constructing} contain fewer than 5\% of aggregation questions. Numerical reasoning datasets (e.g., DROP~\cite{dua-etal-2019-drop}) mainly test number extraction or comparison, rather than identifying entity sets satisfying compositional conditions. Therefore, to develop and evaluate methods for aggregation query over text, it is now urgently required to construct benchmarks that cover different querying scenarios as well as their associated evaluation protocols.


To study this challenge, we formalize a controlled yet representative setting for aggregation querying over text. We focus on entity-level aggregation, where the goal is to identify the complete set of entities in a corpus that satisfy compositional natural-language conditions. This formulation captures the core difficulty of aggregation over unstructured text: grounding conditions and instances under strict completeness requirements.

Based on this formulation, we construct a benchmark called \textbf{\textsc{AGGBench}} for completeness-oriented aggregation over text. Benchmarking this task involves a fundamental trade-off: high-quality ground truth requires expensive evidence grounding, while realistic evaluation demands large, noisy corpora with sparse relevant entities. To reconcile these constraints, we introduce a benchmark construction pipeline that establishes reliable ground truth on a controlled subset and evaluates systems under corpus expansion, enabling scalable and faithful assessment of aggregation querying.

Finally, we propose \textbf{\textsc{DFA} (Disambiguation–Filtering–Aggregation)}, an agentic baseline tailored to entity-level aggregation over text. Motivated by recent advances in general-purpose agentic frameworks~\cite{yao2023react, shen2023hugginggpt, zhou2024languageagenttreesearch}, DFA instantiates an agentic execution paradigm tailored to the specific requirements of aggregation. DFA decomposes aggregation into query disambiguation, completeness-aware filtering, and evidence aggregation, making completeness an explicit and diagnosable objective. Its modular design serves both as a competitive reference system and a foundation for future research.

\noindent\textbf{Contributions:}
\begin{itemize}\setlength\itemsep{0em}
\item We introduce the task of executing aggregation queries over unstructured text at the entity level, formalizing a practically important yet understudied problem that requires exhaustively identifying all entities satisfying compositional natural-language conditions.
\item We present AGGBench, a completeness-oriented benchmark with 362 aggregation queries and evidence-grounded entity annotations. AGGBench scales evaluation to large corpus with extreme evidence sparsity (down to 0.85\%).
\item We propose DFA, a modular agentic framework that operationalizes the key components required for entity-level aggregation querying over text. Across multiple LLM backbones, DFA achieves up to 5× higher evidence recall compared to strong RAG and agentic baselines.
\end{itemize}

%% file: sections/Related_Work.tex
\section{Related Work}
\label{sec:related}

We review benchmarks and methods related to aggregation-style queries over text.

\subsection{Aggregation Query Benchmarks}

Most widely used question answering benchmarks do not explicitly account for aggregation queries or completeness-oriented evaluation. Datasets such as HotpotQA~\cite{yang-etal-2018-hotpotqa} and other multi-hop QA benchmarks~\cite{talmor2018complexwebquestions, trivedi2022musique, khot2020qasc, ho-etal-2020-constructing} identifying a small set of supporting passages, and aggregation-style questions constitute only a small fraction. Similarly, numerical reasoning benchmarks such as DROP~\cite{dua-etal-2019-drop} focus on arithmetic reasoning rather than entity retrieval. More recent benchmarks have begun to target list-based or aggregation-style questions over large text collections \cite{katti2021weblists, liu2023liquid, zhang-etal-2025-aqa, crag2024benchmark, wong2025widesearch}. In contrast, our benchmark is designed to address this limitation by supporting completeness-oriented evaluation and diagnosis of aggregation errors under large retrieval spaces.

\subsection{Methods Related to Aggregation Queries}

Existing approaches relevant to aggregation queries span retrieval-based QA, schema-driven execution pipelines, and agentic LLM frameworks. Retrieval-based QA and RAG systems typically follow a rank-then-read paradigm, in which a retriever returns a fixed number of top-ranked passages and a language model generates an answer conditioned on them \cite{chen-etal-2017-reading, karpukhin-etal-2020-dense, khattab2020colbert, lewis-etal-2020-rag}. Extensions such as multi-hop retrieval, iterative refinement, and retrieval–reasoning interleaving improve coverage in some cases \cite{xiong2021answering, asai2023selfrag, trivedi2023interleaving, mao-etal-2021-generation, su-etal-2024-dragin}, whereas our approach explicitly models completeness and performs iterative retrieval and validation until convergence.

Schema-first pipelines aim to improve executability by mapping text to structured representations and applying symbolic queries, such as in Text-to-SQL systems \cite{zhong2017seq2sql, yu-etal-2018-spider, wang-etal-2020-rat}. Our approach avoids schema construction and operates directly on raw text, allowing aggregation over unstructured corpora with fuzzy or compositional constraints.

More recently, agentic LLM frameworks introduce iterative control flows and tool use for complex reasoning \cite{yao2023react, shen2023hugginggpt, zhou2023webarena, madaan2023selfrefine,  zhou2024languageagenttreesearch}. In this work, we adapt the agentic paradigm to aggregation querying by incorporating design choices motivated by the find-all requirement, including completeness-aware iteration and validation mechanisms.

%% file: sections/Preliminaries.tex
\section{Problem Statement} \label{sec:problem}

We target entity-level aggregation queries over text. Given an aggregation query $q$ and a text corpus $\mathcal{C}$, the goal is to retrieve the complete set of entities that satisfy conditions specified in $q$. Without loss of generality, an aggregation query $q$ can be decomposed into two components: an entity type $T$, and a set of conditions $\Phi = \{\phi_1, \phi_2, \dots, \phi_m\}$. 

\paragraph{Entity type.} $T$ specifies the category of entities to be retrieved. Each query $q$ is assumed to focus on a single entity type. 

\paragraph{Conditions.} Each query specifies a set of conditions $\Phi$ that constrain the target entity set.
We treat each condition $\phi$ as a boolean predicate over a single entity of type $T$, which evaluates to true if and only if there exists explicit textual evidence in the corpus that supports the entity satisfying the condition. These conditions are composed using logical conjunction (AND) and disjunction (OR). More complex logical expressions can be equivalently represented through compositions of these operators (e.g., conjunctive or disjunctive normal form), as commonly assumed in database query processing. Throughout this work, we assume a corpus-bounded setting, in which the corpus $\mathcal{C}$ serves as the sole source of evidence for evaluating all conditions in $q$. The model is therefore prohibited from introducing external knowledge, in contrast to open-domain QA~\cite{karpukhin-etal-2020-dense, wong2025widesearch} settings where external facts may be implicitly assumed.

Let $\mathcal{E}(\mathcal{C})$ denote the set of all entities mentioned in corpus $\mathcal{C}$. The result of an aggregation query $q$ over corpus $\mathcal{C}$ is the set of entities:
\begin{equation}
    \begin{aligned}
    \hspace{0pt}
    \operatorname{Ans}(q, \mathcal{C}) = \{ &e \in \mathcal{E}(\mathcal{C}) \mid \text{type}(e)=T, \\
    &\ \forall \phi_i \in \Phi,\ \phi_i(e)=\text{true} \}.
    \end{aligned}
\end{equation}

In practice, the corpus $\mathcal{C}$ is split into chunks $\{c_1,\dots,c_M\}$ and processed independently.
Let $\operatorname{Ans}(q, c)$ denote the set of entities in chunk $c$ that satisfy query $q$.
The exact answer is obtained by aggregating results across all chunks:
\begin{equation}
\label{eq: normal_ans}
\operatorname{Ans}(q, \mathcal{C}) = \bigcup_{c \in \mathcal{C}} \operatorname{Ans}(q, c),
\end{equation}
where duplicate entities across chunks are merged. Existing approaches can be viewed as simplified approximations of Equation~\eqref{eq: normal_ans}. In Appendix~\ref{appendix:rag-limit}, we formalize these approximations—such as rank–then–read~\cite{chen-etal-2017-reading} and schema-first pipelines~\cite{wang-etal-2020-rat}—and analyze why their design leads to incompleteness.

%% file: sections/Benchmark.tex
\section{Benchmark: AGGBench} \label{sec:benchmark}
This section describes the end-to-end pipeline for constructing \text{AGGBench}, a benchmark for executing aggregation queries over raw text in a corpus-bounded, evidence-grounded setting.

\begin{figure}[t]
  \centering
  \includegraphics[width=\linewidth, trim=0 0 0 0, clip]{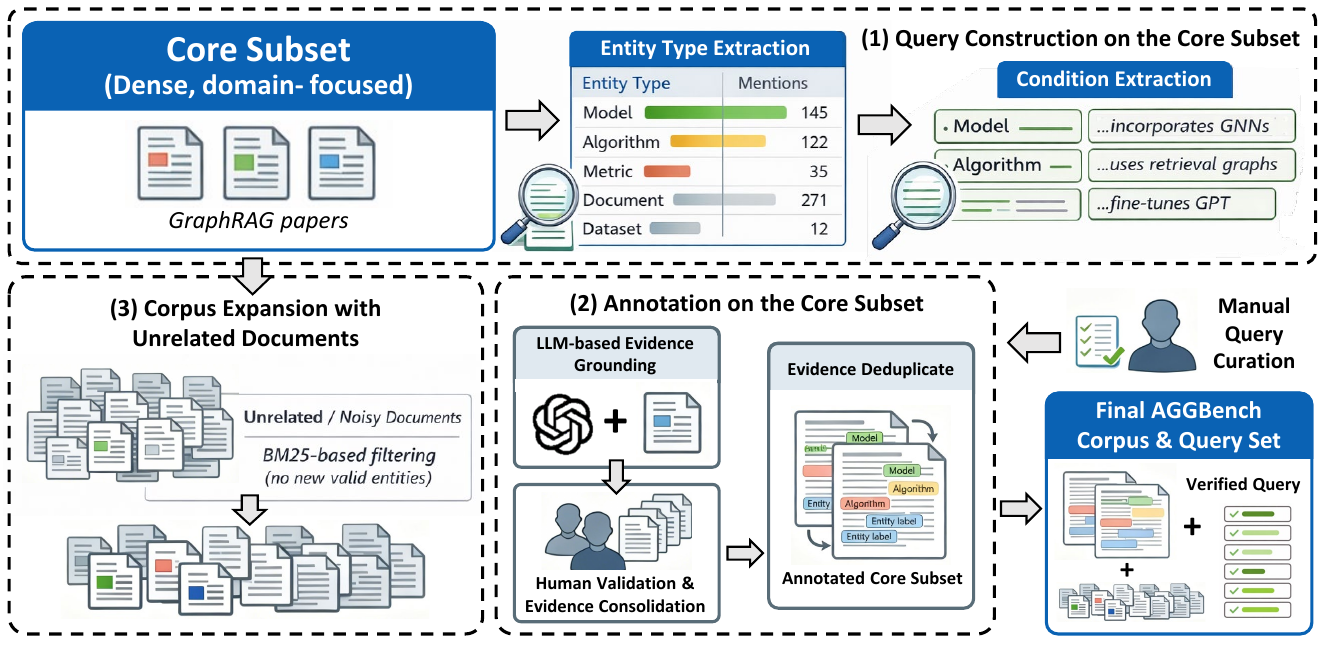}
  \caption{Overview of the AGGBench construction pipeline.}
  \label{fig:pipeline}
  \vspace{-2mm}
\end{figure}

\subsection{Pipeline Overview}
\label{sec:benchmark:pipeline_overview}

We adopt a three-stage construction pipeline, as illustrated in Figure~\ref{fig:pipeline}. Specifically, the pipeline consists of (1) query construction on a controlled core subset, (2) evidence-grounded annotation on the core subset, and (3) corpus expansion with unrelated documents.

\subsection{Query Construction on the Core Subset}
\label{sec:benchmark:corpus}

In the first stage, we construct aggregation queries on a controlled core subset that is domain-focused. We begin by curating a controlled core subset of documents $\mathcal{C}$ that are relevant to graph retrieval-augmented generation. These papers are collected from recent conference and preprint venues and form a domain-focused corpus.

Queries are constructed on the core subset. We first identify a set of frequent entity types $T$ by applying an entity extraction pipeline to the corpus and computing corpus-level mention frequencies, followed by manual filtering to remove overly generic or ill-defined types. For each retained type $T$, we mine high-frequency descriptive phrases and contextual patterns to construct candidate conditions $\phi$. A subset of these conditions is manually curated to ensure semantic unambiguity and satisfiability under the corpus-bounded assumption.

Each query $q$ is then formulated as an aggregation-style natural-language prompt that requires to find all entities satisfying one or more conditions. For example, when $T=dataset$, conditions such as "used for multi-hop question answering" yield queries like: “How many datasets are used for multi-hop question answering?”


When no explicit extractable entity type is available, we adopt a fallback strategy by treating document as the entity type T. In this case, candidate conditions are derived by mining high-frequency concepts or techniques from the corpus. A document is considered to satisfy a query if it contains a grounded mention of the specified concept or technique. This approach yields queries such as “How many documents mention Transformer?”. 

\subsection{Annotation on the Core Subset}
\label{sec:benchmark:annotation}

In the second stage, we annotate the core subset with evidence-grounded labels following Equation~\eqref{eq: normal_ans}, identifying all entities that satisfy each query and grounding them to textual evidence. Fully manual annotation is prohibitively expensive at scale. We therefore adopt a two-stage annotation pipeline. First, an LLM is used to automatically evaluate each query–chunk pair, determining whether the chunk provides evidence that a candidate entity satisfies the query conditions. This step filters out a large number of clear negative cases and produces a reduced candidate set.

Human annotators then verify the remaining cases by confirming or correcting model predictions and consolidating evidence across chunks. Their efforts focus on validating evidence faithfulness, resolving ambiguities, and enforcing consistency at the entity level. In practice, only about 10\% of the LLM-generated labels require revision, indicating that automatic grounding provides a strong initial approximation.

\subsection{Corpus Expansion}
\label{sec:benchmark:expansion}

In the third stage, we expand the corpus by introducing a large collection of unrelated documents, creating a realistic and evidence-sparse retrieval environment while preserving the validity of the annotations. To ensure this, the added documents must not introduce any new entities that satisfy the queries under the evidence-grounded protocol.

We begin by collecting a broad pool of documents that are topically unrelated to the original corpus. To refine this pool, we index the core subset with a BM25 retriever. For each candidate document, we issue it as a query to retrieve the top-k core documents and use the top-1 BM25~\cite{robertson-zaragoza-2009-bm25} score as a proxy for corpus proximity. We then keep documents whose scores fall within an interval, filtering both near-duplicates and clearly out-of-scope documents. After corpus expansion, we also revisit the query set and remove those overly generic entity types to prevent ambiguous or unbounded answers.




%% file: sections/Methodology.tex
\section{DFA Framework}
\label{sec:method}

\begin{figure}[t]
  \centering
  \includegraphics[width=\linewidth, trim=0 0 0 0, clip]{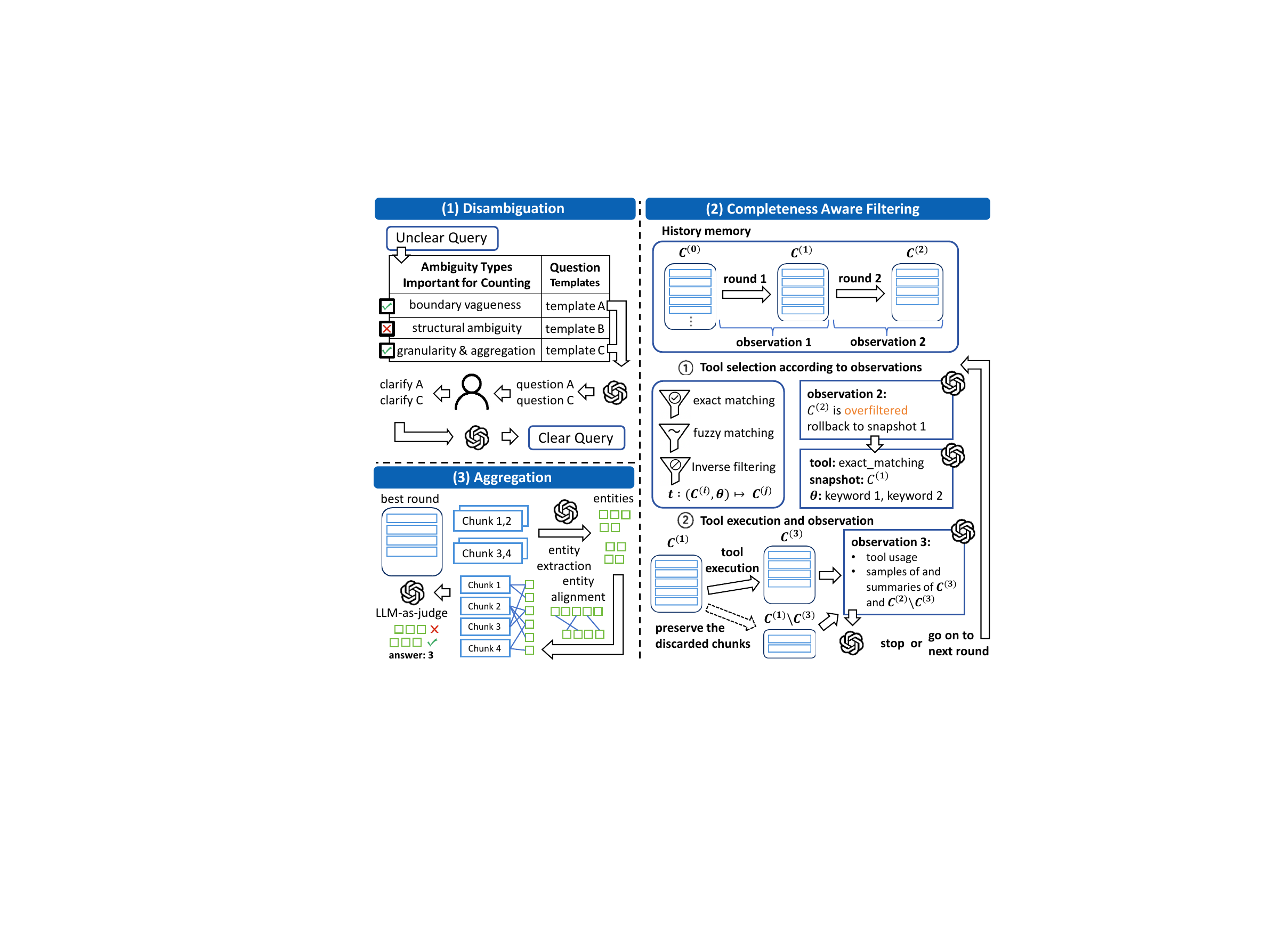}
  \caption{Overview of the DFA framework.}
  \label{fig:dfa}
  \vspace{-2.5mm}
\end{figure}

We propose DFA as a strong baseline for the benchmark. Aggregation queries fundamentally require completeness—identifying all entities that satisfy the query—but achieving this over large unstructured corpora also involves resolving query ambiguities and aggregating evidence across documents, which are not explicitly modeled in existing retrieval or agentic frameworks.

DFA makes these challenges explicit by decomposing entity-level aggregation into three modular stages: disambiguation, completeness-aware filtering, and aggregation (Figure~\ref{fig:dfa}). This design exposes key failure modes and provides a clear reference framework for analyzing and improving completeness-oriented aggregation over text.

\subsection{Disambiguation}
\label{sec:method:disamb}

Disambiguation is a necessary component for executing aggregation queries over text. Under a \emph{find-all} requirement, even minor ambiguities in query interpretation can lead to substantial downstream inefficiency. As a result, both inference cost and latency become highly sensitive to how the query is interpreted.

To address this, DFA introduces a lightweight disambiguation step before any retrieval or filtering is performed. Given a query $q$, the system first identifies potential sources of ambiguity and classifies them into predefined sub-types; details are provided in Appendix~\ref{appendix:ambiguity-subtypes}. It then applies tailored prompt templates to generate clarification questions, and rewrites the query according to the confirmed interpretation before any costly retrieval or aggregation is carried out. This design extends prior interactive clarification approaches~\cite{erbacher-etal-2022-interactive} with a taxonomy tailored to aggregation-oriented querying, where ambiguity directly affects coverage and efficiency.

\paragraph{Insight.}
By isolating disambiguation as a separate module, DFA enables alternative implementations or dedicated ambiguity-resolution models. Future works could also apply this module offline when constructing the benchmark to reduce ambiguity in aggregation benchmarks and improve evaluation stability.

\subsection{Completeness-Aware Filtering}

After disambiguation, DFA performs iterative filtering over the chunked corpus to identify all evidence required to evaluate the query. Unlike relevance-driven retrieval pipelines that optimize for precision, this stage is explicitly designed to preserve recall and avoid irreversible evidence loss.


The baseline implements completeness-aware filtering as an iterative process over explicit corpus snapshots. Starting from an initial snapshot $C^{(0)}$, the system maintains a sequence of corpus states $\{C^{(i)}\}$, each representing the set of remaining candidate chunks after one round of filtering. At each iteration, the system observes the current snapshot $C^{(i)}$ together with the filtering history stored in memory, and selects a filtering tool (e.g., exact matching, fuzzy matching) along with a set of parameters $\theta$. The selected tool is then executed on $C^{(i)}$ to produce a new snapshot $C^{(i+1)}$.

Crucially, all filtering operations are performed on explicit snapshots, and discarded chunks are preserved rather than permanently removed. If the system detects that the current snapshot is over-filtered, it can roll back to an earlier snapshot (e.g., $C^{(1)}$) and reapply a different filtering strategy. Observations, tool usages, and summaries of both retained and discarded chunks are recorded in history memory, enabling informed rollback and iterative refinement across rounds.

\paragraph{Insight.}
DFA adopts a minimal yet explicit filtering design to provide a clear reference point, while exposing several directions for extension. Future works can build more flexible memory across filtering iterations, allowing the system to reason over past filtering decisions rather than treating each step independently. The observation mechanisms over retained and discarded content can also be more adaptive, enabling the model to detect over-filtering signals and adjust its strategy.

\subsection{Aggregation}

The final stage aggregates evidence across candidate chunks to determine which entities satisfy all query conditions. This requires fine-grained verification and alignment, as evidence for the same entity may appear across multiple chunks. 

DFA adopts a batch-based aggregation strategy that applies LLM-as-judge within context limits and aligns extracted entities across batches to remove duplicates; details are provided in Appendix~\ref{appendix:aggregation_alg}. To reduce the computational cost of alignment, the baseline groups semantically similar chunks and processes related content together, minimizing cross-batch comparisons.

\paragraph{Insight.}
Aggregation can naturally be extended beyond a one-pass verification step. Evidence identified at this stage may indicate missing contexts or insufficient support, providing signals for further retrieval or relaxed filtering in targeted regions of the corpus. Future works could couple aggregation and filtering tightly, enabling iterative evidence discovery while maintaining completeness guarantees.

%% file: sections/Experiments.tex
\section{Experiments} \label{sec:expt}

In this section, we conduct experiments on the proposed benchmark to analyze the behavior of different paradigms for aggregation over text.

\subsection{Experiment Setup}\label{subsubsec:experiment_setup}

\begin{table*}[t]
\centering
\footnotesize 
\caption{
Main performance comparison on the AGGBench and AGGBench-Core datasets using GPT-4o-mini. 
The best score in each column is highlighted in \textbf{bold}, while the second-best score is \underline{underlined}.
}
\label{tab:main performance}
\resizebox{\textwidth}{!}{ 
\setlength{\tabcolsep}{1pt}
\renewcommand{\arraystretch}{1.3}
\begin{tabular}{@{}llcccccccc@{}}
\toprule
\multirow{2}{*}{Category} & \multirow{2}{*}{Method} & \multicolumn{4}{c}{AGGBench} & \multicolumn{4}{c}{AGGBench-Core} \\
\cmidrule(lr){3-6}\cmidrule(lr){7-10}
 &  & NACE (mean) & NACE (med) & ACE (mean) & Recall & NACE (mean) & NACE (med) & ACE (mean) & Recall \\
\midrule

    LLM-only & LLM-only & 1.770 & 1.000 & 3.741  & 0  & 1.083 & \underline{0.500} & 3.908  & 0   \\
\midrule

\multirow{3}{*}{Single-Step RAG} 
 & Naive RAG & \underline{0.533} & \underline{0.450} & \textbf{2.327}  & 0.008 & \textbf{0.463} & \underline{0.500} & 3.522  & 0.042   \\
 & LightRAG~\cite{guo2025lightragsimplefastretrievalaugmented}  & 1.311 & 0.542 & 3.028 & 0.031 & 0.902 & \underline{0.500} & 3.461   & \underline{0.096}   \\
 & HippoRAG~\cite{jimenez2024hipporag}     & \textbf{0.456} & \textbf{0.414} & \underline{2.351}   & \underline{0.092}   & \underline{0.503} & \underline{0.500} & \underline{3.511}   & 0.041   \\
\midrule

\multirow{2}{*}{Multi-Hop RAG} 
 & DeepNote~\cite{wang2025deepnotenotecentricdeepretrievalaugmented}     & 4.174 & 0.667 & 10.760   & 0.031  & 0.621 & 0.667 & 4.552   & 0.015   \\
 & RAT~\cite{wang2024ratretrievalaugmentedthoughts}          & 0.793 & 0.500 & 2.670   & 0.018   & 0.787 & \underline{0.500} & 4.076   & 0.038   \\
\midrule

\multirow{4}{*}{Agentic Framework} 
& ReAct~\cite{yao2023react} & 0.807 & 0.750 & 3.463   & 0.030   & 0.770 & \underline{0.500} & 3.773   & 0.054   \\
& Single Agent~\cite{wong2025widesearch} & 1.966 & 0.733 & 6.574   & 0.042   & 1.522 & \underline{0.500} & 5.939   & 0.031   \\
& Multi-Agent~\cite{wong2025widesearch} & 1.891 & 1.000 & 5.500  & 0.044   & 0.770 & 0.571 & 6.412  & 0.026   \\
& smolagents~\cite{smolagents} & 0.742 & 0.573 & 2.667 & 0.026  & 1.079 & \underline{0.500} & 6.820  & 0.050   \\
\midrule

    \textbf{Ours} & DFA & 0.861  &   0.667 & 3.174   &\textbf{0.211}    & 0.544  & \textbf{0.464}   & \textbf{2.670}   & \textbf{0.623}   \\
\bottomrule
\end{tabular}
}
\vspace{-12pt}
\label{tab:method_comparison}
\end{table*}

\noindent\textbf{Datasets.} We conduct our evaluation on three benchmarks, all equipped with ground-truth evidence annotations. The primary benchmark, AGGBench, is constructed following the pipeline described in Section~\ref{sec:benchmark}. We also evaluate on a variant of AGGBench without corpus expansion, called AGGBench-Core. It includes more queries and allows isolating aggregation performance under a controlled and less noisy retrieval setting. As a complementary benchmark, we derive \textbf{SpiderQA} from the Text-to-SQL dataset \textbf{Spider}~\cite{yu-etal-2018-spider} by extracting questions involving aggregation operators (e.g., COUNT) and converting their structured corpus to unstructured text. In addition, we introduce supplementary datasets to separately evaluate the modules for disambiguation and aggregation. Details are provided in Appendix~\ref{appendix:dataset}. 

\noindent\textbf{Reference Methods.} We evaluate a diverse set of representative paradigms on our benchmark, including DFA as a strong baseline, as shown in Table~\ref{tab:method_comparison} and Appendix~\ref{appendix:detail_performance}. The LLM-only baseline directly prompts LLM without retrieval. The Naive RAG baseline uses standard top-$k$ retrieval with chunk-based RAG prompting. We also compare our method with the three categories of Single-Step RAG, Multi-Hop RAG, and Agentic Framework. More details are provided in the Appendix~\ref{appendix:baseline}.

\noindent\textbf{LLM Models.} The experiments were conducted on four representative LLMs—GPT-4o-mini, Gemini-3-Flash, DeepSeek-R1, and Qwen3-32B. They cover both large and small models, open-source and proprietary settings, as well as reasoning and non-reasoning variants.

\noindent\textbf{Metrics and Implementation.}
For aggregation queries, we evaluate execution results along two essential dimensions: \textbf{evidence completeness} and \textbf{result accuracy}.
Evidence completeness is measured using chunk-level \textbf{recall}, defined as the proportion of gold evidence chunks that are retained in the filtered candidate set after retrieval and filtering. Result accuracy is assessed using \textbf{Absolute Count Error (ACE)} and \textbf{Normalized Absolute Count Error (NACE)}, capturing the absolute error in the predicted number of entities that satisfy the query conditions. ACE is defined as $\text{ACE} = |\hat{y} - y|$, measuring the absolute deviation between the predicted and ground-truth counts. NACE is defined as: $\text{NACE} = \frac{|\hat{y} - y|}{y + \epsilon}$ where $\hat{y}$ denotes the size of the predicted entity set, $y$ is the ground-truth size, and $\epsilon$ is a small constant to prevent division by zero. Note that NACE values can exceed~1 when the predicted count deviates from the ground truth by more than the true value itself. More implementation details are provided in Appendix~\ref{appendix:impl}.

\subsection{Benchmark Characterization} \label{sec:benchmark_characterization}

We first characterize the proposed benchmark from both corpus- and query-level perspectives. 

\subsubsection{Corpus and Query Statistics}
\label{sec:corpus_query_stats}

We summarize statistics of the benchmark corpus and queries. At the corpus scale, our benchmark is built over 45 related research papers and contains 362 aggregation queries. Among these, 100 are base aggregation queries with a single condition, while the remaining 262 are composite queries formed by set-level union or intersection over the base queries.

At the query level, we analyze the distribution of aggregation queries by the number of valid entities satisfying the query conditions. Among the 362 queries, 165 have answer sizes greater than 5, showing coverage of both small-count and medium-to-large aggregation scenarios. The maximum answer size is 20 for single-condition queries and 29 for composite queries. Importantly, answering such queries requires retrieving a corresponding number of distinct evidence mentions from the corpus. This evidence requirement is substantially higher than that of conventional question answering benchmarks, where queries typically involve only one to a few supporting facts.

Regarding query structure, the 262 composite queries in AGGBench include 42 conjunctive (AND) queries and 220 disjunctive (OR) queries. Most composite queries involve two conditions (228 queries), while a smaller subset involves three conditions (34 queries), introducing higher-order combinatorial complexity. This distribution shows that the benchmark mainly emphasizes union-style aggregation while maintaining a smaller set of more complex compositions.

\subsubsection{Evidence Sparsity Analysis}
\label{sec:evidence_sparsity}

One of the challenges of aggregation over large text corpora is evidence sparsity: the evidence required to answer a query typically constitutes only a small fraction of the entire corpus. To quantify this effect, we analyze evidence sparsity under corpus expansion. In the base corpus, the collection contains 4,755 text chunks, among which 294 are directly relevant evidence chunks, yielding an evidence density of 6.18\%. In contrast, under the expanded corpus setting, the corpus grows to 16,294 text chunks, while the number of directly relevant evidence chunks decreases to 178 due to the reduced number of queries, resulting in a substantially lower evidence density of only 1.09\%.


\subsection{Performance Landscape on AGGBench} \label{sec:main_results}
We evaluate the proposed benchmark across a range of representative paradigms and language models to map out the performance landscape of aggregation query execution.
We further conduct an in-depth analysis of DFA as a strong baseline.

\subsubsection{Performance of Existing Paradigms}

The results are shown in Table~\ref{tab:main performance}. Overall, existing paradigms struggle to reliably execute aggregation queries under evidence-sparse and noisy retrieval settings. We analyze reasons for their failure.

\noindent\textbf{Single-step RAG.} Single-step RAG methods perform poorly across both benchmarks, exhibiting near-zero recall in most cases. Even schema-first approaches that incorporate structured retrieval signals fail to substantially improve performance. This is primarily because they are highly sensitive to noisy chunks. The core issue lies in aggregation tasks: the semantic gap between natural-language queries and the corresponding evidence is often large, making direct retrieval using the original query formulation ineffective.

\noindent\textbf{Multi-hop RAG.} In terms of recall, multi-hop approaches improve from near-zero levels (typically below 0.01–0.04) to approximately 0.018–0.031, corresponding to roughly 2–5× relative gains depending on the benchmark. This improvement stems from their ability to iteratively refine the query. However, these methods are primarily designed for "deep" reasoning, where the goal is to identify a small set of highly relevant evidence. In contrast, aggregation requires "wide" retrieval to ensure completeness.

\noindent\textbf{Agentic frameworks.} Agentic frameworks achieve better overall performance than RAG-based approaches, reflecting their stronger control over retrieval and reasoning processes. Their recall scores (approximately 0.026–0.044 across settings) consistently exceed those of single-step and multi-hop RAG. However, these methods lack explicit mechanisms to track or reason about evidence discarded in earlier retrieval and filtering steps, making them prone to over-filtering and the omission of partially relevant evidence in aggregation settings.

\subsubsection{Performance of DFA}
DFA performs best on AGGBench-Core, achieving the lowest errors and the highest recall, outperforming the strongest baseline recall by over 5×. It suggests that the improved accuracy is driven by comprehensive evidence collection rather than heuristic guessing or over-generation. On the full AGGBench, DFA attains the highest recall, but its errors are not the best. It suggests future directions to focus on improving robustness to larger corpus. We also analyze the design choices of DFA by examining each execution module, with more details presented in Appendix~\ref{sec:ablation}.

\subsubsection{Comparison Across Base Models}

\begin{table}[t]
\centering
\caption{
Main performance comparison on AGGBench across base models. 
}
\label{tab:model_performance_comparison}
\resizebox{0.47\textwidth}{!}{ 
\setlength{\tabcolsep}{3pt} 
\renewcommand{\arraystretch}{1.0} 
\begin{tabular}{@{}lcccc@{}} 
\toprule
Model & NACE (mean) & NACE (med) & ACE (mean) & Recall \\
\midrule
GPT-4o mini    & 0.861  &   0.667 & 3.174   &0.211 \\
Gemini-3-Flash & 0.732   & 0.750   & 2.870   & 0.490   \\
Qwen-3 32b     & 1.733  & 0.833  & 5.065   & 0.482   \\
DeepSeek-R1    & 0.950   & 1.000   & 4.000   & 0.378   \\
\bottomrule
\end{tabular}
}
\vspace{0pt}
\end{table}

We evaluate our framework on AGGBench across a diverse set of LLMs to assess its robustness and model-agnostic behavior. As shown in Table~\ref{tab:model_performance_comparison}, Gemini-3-Flash achieves the best overall performance and outperforms all other models, which is consistent with prior reports that Gemini-3 has been explicitly trained to support broad and exhaustive information-seeking and search-oriented tasks. In contrast, GPT-4o-mini and DeepSeek-R1 exhibit lower performance, reflecting the limitations of earlier-generation proprietary models in handling completeness-oriented retrieval and aggregation. Notably, smaller models and open-source models show only a modest performance degradation, indicating that our framework remains stable across model scales.

\subsection{Diagnostic Analysis}
In this section, we analyze the behaviors and limitations of DFA and other reference methods in executing aggregation queries.

\subsubsection{Failure Modes of Existing Paradigms}
We analyze a representative aggregation query: “How many papers apply to the legal domain?”
NaiveRAG fails on this query because it relies solely on semantic similarity ranking without enforcing domain constraints; the semantic gap between the query and one of the evidence chunks causes the chunk to be ranked low and missed. Further analysis shows that increasing the reasoning trajectory length (RAT) does not alleviate this issue. The system terminates retrieval once partial evidence is found, lacking completeness-aware behavior required for aggregation tasks.

Schema-first methods (e.g., HippoRAG) fail for a different reason: the concept of the legal domain lies outside the predefined schema used during structured corpus construction, preventing relevant evidence from being retrieved. Moreover, these approaches require costly upfront structuring of the corpus, making them brittle to domain shifts.

In contrast, DFA explicitly prioritizes evidence coverage over relevance ranking. It first applies constraint-driven keyword filtering (e.g., legal, law), narrowing the search space, and then dynamically relaxes or tightens matching criteria to balance recall and noise. This coverage-aware, tool-driven strategy enables DFA to recover all ground-truth evidence and produce the correct count.

\subsubsection{Open Challenges and Limitations}
We analyze aggregation queries that DFA answers incorrectly and identify recurring failure patterns that motivate future improvements. About 15\% of errors come from incorrect tool parameterization: the model misuses filtering or matching tools due to limited familiarity with their interfaces. This indicates that DFA assumes static tool competence; incorporating tool-awareness or tool-evolution learning could improve parameter selection and usage reliability. Another approximately 5\% of failures are redundant repeated calls to the same tool across iterations, likely driven by overly long or poorly managed intermediate memory states, motivating adaptive and explicit memory management or compression strategies for multi-round tool execution.

%% file: sections/Conclusion.tex
\section{Conclusion} \label{sec:conclusion}
This work studies aggregation querying over unstructured text as a completeness-oriented problem distinct from conventional question answering. We introduce AGGBench, a benchmark that enables principled evaluation of “find-all” aggregation under realistic, evidence-sparse corpora. We further present DFA, a modular agentic baseline that makes ambiguity, filtering, and aggregation failures explicit and diagnosable. Our findings highlight aggregation as a promising testbed for developing completeness-aware retrieval, memory, and agentic execution strategies. We hope this work will inspire subsequent studies to explore broader applications of completeness-aware agents.

%% file: sections/limitations.tex
\section{Limitations} \label{sec:limitations}
While the DFA framework demonstrates strong performance across multiple real-world datasets, several limitations remain. While our work introduces a modular and interpretable agent-style pipeline, which naturally offers valuable insights into memory management and tool design for broader agent systems, we do not explore these directions in depth due to space constraints. We leave a more thorough investigation of learnable memory modules, dynamic tool orchestration, and planning-aware extensions of DFA to future work.

%% file: sections/ethical_impact.tex
\section{Ethical Impact}
All datasets in this study are constructed from publicly available records and papers.
We do \emph{not} use any data that could directly track persons. Therefore, the typical privacy risks associated with personal searching data do not arise in our setting, and our use complies with community ethics guidelines and licenses.

%% file: sections/appendix.tex
\clearpage
\newpage
\section{Appendix}
\subsection{Limitations of Existing Pipelines} \label{appendix:rag-limit}
Here we formalize previous methods and discuss their limitations. We use $\operatorname{count}(q)$ as shorthand for the cardinality of the exact answer set, i.e.,
$\operatorname{count}(q) = |\mathrm{Ans}(q, \mathcal{C})|$, where $\mathrm{Ans}(q, \mathcal{C})$
denotes the set of entities in corpus $\mathcal{C}$ that satisfy query $q$. Directly computing the expression in Equation~\eqref{eq: normal_ans} is infeasible when $|\mathcal{C}|$ is large, so some existing systems adopt a rank–then–read strategy. A lightweight retriever assigns each chunk a relevance score and keeps only the top $k$ chunks. These $k$ chunks are fed to a large language model $p_{\theta}$.  
When $k\ll\operatorname{count}(q)$, many relevant entities never reach the model, and the resulting upper bound:  
\begin{align}
\widehat{\operatorname{count}}_{\text{RAG}}(q)=\bigl|\operatorname{AGG}_{c\in\mathcal{Z}_k(q, \mathcal{C})}\psi_q(c)\bigr|
\end{align}
can fall far below the true total. Otherwise, increasing $k$ indiscriminately introduces large amounts of irrelevant text, inflates context length, and quickly exceeds the input limit of $p_{\theta}$.

A second line of work replaces free-text reasoning with \textit{schema-first} pipelines.  
Each chunk is pre-processed by an extractor $T$ that maps text to structured tuples  $T:\mathcal{C}\rightarrow\mathcal{S}$, yielding a relational or graph database $\mathcal{S}$.  Then, the original query is rewritten into SQL or SPARQL form, denoted $q^{\star}$, and executed as  
\begin{align}
\operatorname{count}(q)=\bigl|\sigma_{q^{\star}}(\mathcal{S})\bigr|,
\end{align}
where $\sigma_{q^{\star}}(\cdot)$ returns all tuples satisfying the structured query.  
This approach suffers from two issues: building $\mathcal{S}$ demands expensive extraction and alignment, and the fixed schema restricts $q^{\star}$ to explicit attributes, making it hard to capture implicit or compositional conditions.

In short, rank–then–read RAG sacrifices recall for efficiency, while schema-first pipelines incur heavy upfront costs yet remain rigid. Neither class can deliver complete aggregation at a practical inference cost; hence, a new framework is required to solve aggregation problems effectively.

\subsection{Ambiguity Subtypes in Aggregation Queries} \label{appendix:ambiguity-subtypes}

\begin{table*}[t]
\caption{Seven ambiguity sub-types that can distort aggregation. Each query admits two plausible readings leading to markedly different tallies.}
\centering
\scriptsize
\setlength{\tabcolsep}{4pt} 
\begin{tabular*}{\textwidth}{@{\extracolsep{\fill}}p{2.5cm} p{3.5cm} p{9.0cm}}
\toprule
\textbf{Sub-type} & \textbf{Definition} & \textbf{Example (query $\rightarrow$ two readings)} \\
\midrule
A1 Scalar / implicit threshold & Gradable adjective, no cut-off. & Q: How many \emph{high-impact} NLP authors? (a) $h$-index $\ge 50$ $\rightarrow$ 11 authors. (b) Top 1\,\% citations $\rightarrow$ 43 authors. \\
\midrule
A2 Temporal / spatial window & Relative time or distance phrase; window unknown. & Q: How many stores opened \emph{near HQ} \emph{recently}? (a) 5\,km \& last 6 months $\rightarrow$ 8 stores. (b) 20\,km \& last 3 years $\rightarrow$ 47 stores. \\
\midrule
B1 Logical scope & Negation / quantifier scope ambiguous. & Q: How many employees \emph{did not} complete \emph{at least three} safety-training courses? (a) Completed $<3$ courses $\rightarrow$ 189. (b) Not finish three distinct courses $\rightarrow$ 37. \\
\midrule
B2 Attachment ambiguity & Modifier may attach to different heads. & Q: How many conference papers \emph{on climate change} \emph{in Europe} were accepted? (a) Authors located in Europe $\rightarrow$ 18. (b) Topic = European climate change $\rightarrow$ 64. \\
\midrule
C1 Entity-type granularity & Unit can be paper / section / run. & Q: How many \emph{experiments} mention dataset X? (a) Paper level $\rightarrow$ 52. (b) Section level $\rightarrow$ 136. \\
\midrule
C2 Cross-document dedup & Alias merging rule unclear. & Q: Distinct chemical compounds discovered this decade. (a) Merge by InChI Key $\rightarrow$ 7\,900. (b) String match only $\rightarrow$ 10\,200. \\
\midrule
C3 Unknown entity label & Rare label; boundary undefined. & Q: How many official songs from \emph{Ave Mujica} were released in 2024? The model lacks knowledge that “Ave~Mujica” is a 2024 anime; an authoritative song list is required. \\
\bottomrule
\end{tabular*}
\label{tab:counting-ambiguities-full}
\end{table*}

We provide detailed explanations of the ambiguity subtypes introduced in Section~\ref{sec:method:disamb}, 
including the following sub-types: A1 (Scalar / implicit threshold), A2 (Temporal / spatial window), 
B1 (Logical scope), B2 (Attachment ambiguity), C1 (Entity-type granularity), 
C2 (Cross-document deduplication), and C3 (Unknown entity label). Their definitions and examples are shown in Table~\ref{tab:counting-ambiguities-full}.

\paragraph{Boundary vagueness (A-types).}  
A-type ambiguities arise from unclear boundaries within individual conditions, such as gradable adjectives (e.g., “high-impact,” “large”) or relative temporal/spatial phrases (e.g., “recent,” “near headquarters”). These expressions lack explicit numeric thresholds or ranges, leaving the condition open to multiple interpretations.

\paragraph{Structural ambiguity (B-types).}  
B-type ambiguities concern uncertain logical relationships among conditions. Negation, quantifier scope, or modifier attachment can be parsed in different ways, changing how the conditions combine and altering the effective constraints on the aggregation result.

\paragraph{Granularity and aggregation (C-types).}  
C-type ambiguities involve uncertain definitions of the target entity set. They include unclear aggregation units (e.g., paper, section, or experiment run), cross-document deduplication rules for entities with aliases, and rare or undefined entity labels that lack a precise boundary.


\begin{algorithm*} [!t]
\caption{Greedy Cluster Batching for DFA}
\label{alg:agg}
\small
\begin{algorithmic}[1]
\Require Clusters $K$, Max context $M$, Token counter $T$, Cluster centroids $\{\mu_K\}$, Weight $\lambda$
\Ensure Batches $B = \{B_1, B_2, \ldots\}$
\State $B \gets \emptyset$
\For{each $k \in K$}
    \If{$T(k) > M$} \Comment{Large cluster: Split}
        \State Split $k$ into $\{k^{(1)}, k^{(2)}, \ldots\}$ where $T(k^{(r)}) \le max\_ctx$
        \For{each $k^{(r)}$}
            \State $B \gets B \cup \{\{k^{(r)}\}, \mu_k, T(k^{(r)})\}$
        \EndFor
    \Else \Comment{Small cluster: Merge}
        \State $best\_idx, best\_score \gets None, -\infty$
        \For{each $B_q \in B$}
            \If{$T(B_q) + T(k) \le M$}
                \State $best\_score, best\_idx \gets find(\mu_k, \bar{\mu}(B_q), T(B_q), T(k), M)$
            \EndIf
        \EndFor
        \If{$best\_idx \neq None$} \Comment{Merge into best batch}
            \State $\bar{\mu}(B_q) \gets weighted\_avg(\bar{\mu}(B_q), \mu_{K_k}, T(B_q), T(K_k))$
            \State $T(B_q) \gets T(B_q) + T(K_k)$; \quad $B_q \gets B_q \cup \{K_k\}$
        \Else \Comment{Create new batch}
            \State $B \gets B \cup \{\{K_k\}, \mu_{K_k}, T(K_k)\}$
        \EndIf
    \EndIf
\EndFor
\Statex \hspace{-\algorithmicindent}\textbf{Return: } $B$
\end{algorithmic}
\end{algorithm*}

\subsection{Greedy Semantic-Aware Batching Algorithm} \label{appendix:aggregation_alg}
Since the number of candidate chunks produced by filtering stage often exceeds the LLM context limit, we adopt a greedy semantic-aware batching algorithm, as shown in algorithm~\ref{alg:agg}, our algorithm first splits large clusters (\textit{lines 3–6}): when the token count of a cluster exceeds the LLM’s context limit, it is divided into several batches to avoid overflow. For small clusters, a greedy merging strategy is adopted. For a candidate cluster $\mathcal{K}$ and an existing batch $\mathcal{B}$, the combined score function is defined (\textit{line 12}):
\begin{equation}
\begin{aligned}
S(K,\mathcal{B}) 
&= \lambda\,\cos\!\big(\mu_K, \bar{\mu}(\mathcal{B})\big) \\
&\quad + (1-\lambda)\,\frac{T(\mathcal{B})+T(K)}{L_{\mathrm{etx}}}, \lambda\in[0,1]
\end{aligned}
\end{equation}
where $\mu_K$ denotes the cluster centroid, $\bar{\mu}(\mathcal{B})$ the current centroid of the batch, $\cos(\cdot)$ the cosine similarity, $T(\cdot)$ the token count, and $\lambda$ the weight. The first part captures entity similarity, while the second captures space utilization. Once the optimal batch is identified, the small cluster is merged into it, and the batch centroid is updated by token-weighted averaging to preserve the validity of subsequent similarity computations (\textit{lines 13–15}):

\begin{equation}
\bar{\boldsymbol{\mu}}(\mathcal{B}\cup K)
= \tfrac{T(\mathcal{B})}{T(\mathcal{B})+T(K)}\,\bar{\boldsymbol{\mu}}(\mathcal{B})
+ \tfrac{T(K)}{T(\mathcal{B})+T(K)}\,\mu_K
\end{equation}

If no batch can accommodate the cluster (\textit{lines 16-17}), a new batch is created directly.

\subsection{Baseline} \label{appendix:baseline}
We provide detailed descriptions of the baseline methods introduced in Section~\ref{subsubsec:experiment_setup}, covering four categories and six representative methods:

\begin{itemize}
    \item \textbf{Naive RAG.} A simple retrieval-augmented setup using standard top-$k$ retrieval combined with chunk-based RAG prompting.
    \item \textbf{Single-Step RAG.} Includes improved RAG baselines that apply graph-based retrieval and top-$k$ chunk selection strategies, such as HippoRAG~\cite{jimenez2024hipporag} and LightRAG~\cite{guo2025lightragsimplefastretrievalaugmented}.
    \item \textbf{Multi-Hop RAG.} Covers state-of-the-art multi-hop RAG methods that iteratively retrieve and rewrite queries, including DeepNote~\cite{wang2025deepnotenotecentricdeepretrievalaugmented} and RAT~\cite{wang2024ratretrievalaugmentedthoughts}.
    \item \textbf{Agentic Framework.} This category covers general-purpose agentic framework with retrieval tools. We include ReAct~\cite{yao2023react}, tool-centric Single Agent and Multi-Agent~\cite{wong2025widesearch}, as well as code-oriented agents smolagents~\cite{smolagents}.
\end{itemize}

We follow the descriptions in these works to preprocess the data into the corresponding format and run the baselines with the best parameter settings to ensure fairness.

\subsection{Dataset Construction} \label{appendix:dataset}

we derive \textbf{SpiderQA} benchmark from the Text-to-SQL dataset \textbf{Spider}~\cite{yu-etal-2018-spider}. Supplementary datasets for ambiguity resolution and aggregation module evaluation are also introduced in details.

\subsubsection{Converting Text-to-SQL dataset}
Text-to-SQL benchmarks contain a substantial portion of aggregation-related questions, yet their data is organized in structured database form rather than natural language text. To align with the goal of aggregation in text corpus, we transform a representative Text-to-SQL dataset into text-based documents. We adopt the Spider benchmark, which consists of 166 databases and over 9,000 questions, as the source dataset for conversion.

For each database, we extract its tables' schema information, including table names, column names, primary keys, and foreign keys. Given this structural information, we prompt an LLM to generate a textual template describing each table. Then, the actual table entries are retrieved using SQL queries and filled into the corresponding templates, resulting in a document representation of the entire database.

For questions, we process them by filtering out non-aggregation types (e.g., enumeration, comparison, or “most-value” questions), retaining only those that explicitly involve aggregation. For each remaining question, we execute both the original SQL query and a slightly modified version on the target database to obtain the final answer and supporting evidence, respectively. These question–answer–evidence triples are then attached to their corresponding converted database documents.

To ensure dataset quality and balance, we exclude overly large databases or those with few aggregation questions by computing the ratio between database size and question count, selecting moderate cases to maintain both textual manageability and question diversity. We further remove duplicated tables and repeated questions to guarantee uniqueness and consistency.
The resulting benchmark, termed \textbf{SpiderQA}, comprises 48 text-based database documents and 201 manually verified aggregation question–answer–evidence triples.

\begin{table}[t]
\centering
\caption{Ambiguity classification results on the ACQ dataset measuring Accuracy (Acc.) and F1.}
\label{tab:classification}
\scriptsize
\setlength{\tabcolsep}{2.5pt} 
\begin{tabular}{lcccccccc}
\toprule
\multirow{2}{*}{\textbf{Model}} &
\multicolumn{2}{c}{\textbf{A1}} &
\multicolumn{2}{c}{\textbf{A2}} &
\multicolumn{2}{c}{\textbf{B1}} &
\multicolumn{2}{c}{\textbf{B2}} \\
\cmidrule(lr){2-3} \cmidrule(lr){4-5} \cmidrule(lr){6-7} \cmidrule(lr){8-9}
 & Acc. & F1 & Acc. & F1 & Acc. & F1 & Acc. & F1 \\
\midrule
Gemini-2.5-Flash & 95.88 & 82.95 & 92.14 & 84.59 & 28.83 & 44.76 & 57.97 & 72.73 \\
GPT-4o mini          & 91.95 & 90.40 & 96.18 & 91.97 & 65.43 & 79.10 & 52.94 & 68.57 \\
DeepSeek-R1      & 96.47 & 80.20 & 89.29 & 86.81 & 84.66 & 91.69 & 59.42 & 68.91 \\
Qwen3 32B        & 95.29 & 85.94 & 87.86 & 88.49 & 68.71 & 81.16 & 34.78 & 50.00 \\
\bottomrule
\end{tabular}
\vspace{-7pt}
\end{table}

\subsubsection{Ambiguous Aggregate Questions Dataset}
For the ambiguity resolution module, existing benchmarks on ambiguous queries provide limited coverage of aggregation problems. To fill this gap, we construct an Ambiguous Aggregation Questions dataset. Candidate questions were first generated using GPT-4, along with potential explanations of their ambiguity types. These were subsequently validated and labeled by human annotators, resulting in a curated benchmark of 492 items. Each question is annotated with one or more ambiguity categories from {A1, A2, B1, B2}, covering both single-source and intertwined ambiguity cases. Category~C was excluded from large-scale evaluation due to its relative rarity; instead, representative C-type cases are examined qualitatively.

\subsubsection{Datasets for Aggregation Evaluation}
For evaluating the aggregation module, we use the \textbf{AGGBench-Core} and \textbf{MusiQue} datasets. For MusiQue, 3059 chunks are randomly sampled, and 30\% of the chunks from both datasets are used to simulate the outputs of the filtering module, ensuring consistent and controlled aggregation experiments.

\subsection{Detailed Performance of DFA Framework}
\label{appendix:detail_performance}

We report the detailed performance of the DFA, including its overall performance on the SpiderQA benchmark, the effectiveness of the disambiguation module, and the efficiency gains of the aggregation module.

\subsubsection{Performance Landscape on SpiderQA} \label{appendix:spiderqa_method_comparison}

We evaluate the SpiderQA benchmark, which is introduced in Appendix~\ref{appendix:dataset}, across a range of representative paradigms using GPT-4o-mini. The results are presented in Table~\ref{tab:spiderqa_method_comparison}.

\begin{table}[t]
\centering
\caption{
Main performance comparison on the SpiderQA dataset using GPT-4o-mini. 
The best score in each column is highlighted in \textbf{bold}, while the second-best score is \underline{underlined}.
}
\label{tab:spiderqa_method_comparison}
\resizebox{1.0\columnwidth}{!}{ 
\setlength{\tabcolsep}{3pt} 
\renewcommand{\arraystretch}{1.0} 
\small 
\begin{tabular}{@{}lcccc@{}}  
\toprule
Method & NACE (mean) & NACE (med) & ACE (mean) & Recall \\
\midrule
LLM-only & 5.494   & 0.600   & 45.940   & 0   \\
\midrule
Naive RAG & 0.666   & 0.800   & 9.895   & 0.181   \\
LightRAG     & 0.708   & 0.750   & 9.562   & \underline{0.865}   \\
HippoRAG     & 0.454   & 0.2  & 6.522   & 0.827   \\
\midrule
DeepNote     & 0.425   & 0.162   & \underline{6.495}   & 0.827   \\
RAT          & 0.385   & 0.250   & 7.150   & 0.736   \\
\midrule
ReAct & 0.672   & \underline{0.125}   & 9.529 & 0.614   \\
Single Agent & 0.429   & 0.235   & 6.547 & 0.868   \\
Multi-Agent & 0.632   & 0.222   & 7.761 & 0.856   \\
smolagents & \underline{0.355}   & \textbf{0.100}   & 6.587 & 0.806   \\
\midrule
DFA (Ours) & \textbf{0.262}   & 0.143   & \textbf{4.000}   & \textbf{0.891}   \\
\bottomrule
\end{tabular}
}
\vspace{0pt}
\end{table}

\subsubsection{Performance of Disambiguation Module}

We analyze how well the disambiguation module handles different types of ambiguous queries. Here we focus on ambiguity classification. For each question, the model predicts a set of ambiguity labels from \{A1, A2, B1, B2\}, and a prediction is deemed correct if the predicted set fully covers the gold annotations. Table~\ref{tab:classification} shows that models achieve high accuracy overall. A-type ambiguities (A1/A2) yield higher accuracy than B-type (B1/B2), indicating that surface-level ambiguities are easier to detect than logical ones. This also suggests that smaller models can accomplish the disambiguation task more effectively when guided by templates.

\subsubsection{Performance of Aggregation Module}

\begin{table}[t]
\centering
\caption{%
Aggregation results with different clustering methods. 
Reported by the number of LLM calls (\textbf{LC}) and candidate pairs (\textbf{CP}) on AGGBench-Core and Musique.
}
\label{tab:aggregation}
\resizebox{0.34\textwidth}{!}{
\begin{tabular}{lcccc}
\toprule
\multirow{2}{*}{\textbf{Method}} 
 & \multicolumn{2}{c}{\textbf{AGG-Core}} 
 & \multicolumn{2}{c}{\textbf{MusiQue}} \\
\cmidrule(lr){2-3} \cmidrule(lr){4-5}
 & LC & CP & LC & CP \\
\midrule
GMM      & \textbf{5} & 93  & \underline{26} & 965  \\
Leiden   & 7 & \underline{63}  & 33 & \underline{799}  \\
K-Means  & 7 & 70  & 28 & 917  \\
\textbf{Ours} & \textbf{5} & \textbf{41} & \textbf{25} & \textbf{569} \\
\bottomrule
\end{tabular}
}
\vspace{-7pt}
\end{table}

We isolate the aggregation module by assuming the same set of candidate chunks and compare different clustering strategies, including GMM~\cite{dempster1977maximum}, Leiden~\cite{traag2019louvain}, and K-Means~\cite{lloyd1982least}. We evaluate the number of LLM calls (LC) and the number of candidate pairs that need to be compared across batches (CP). Our method reduces LLM calls by roughly 4\%, indicating that the proposed greedy batching strategy effectively improves batch utilization without increasing inference cost. Moreover, our method achieves a reduction of about 47\% in candidate pairs, demonstrating that our clustering effectively groups semantically similar chunks—those containing entities likely to align during extraction—into the same batches. Among the baselines, Leiden and ours perform noticeably better because they capture hierarchical similarity, leading to more coherent batch formation.

\subsection{Implementation details} \label{appendix:impl}

We specify the implementation details of the experiments, including unified corpus and chunking configurations for all baselines, specific parameter settings of different retrievers, as well as the embedding and reasoning models adopted throughout the study.

\subsubsection{Retrieval settings}
We evaluate multiple retrieval-based baselines under consistent corpus and chunking configurations. 

For baselines that rely on a vector database, all retrievers operate on the same vector store constructed from the text corpus, where passages are embedded using the chosen embedding model and stored as dense vectors. 
Cosine similarity is used for ranking unless otherwise specified.

For NaiveRAG baseline, we employ a vector-based retriever with top-$k \in \{5,10,20,50\}$. 
LightRAG is configured in hybrid retrieval mode, combining vector and keyword matching. The entity top-$k$ is set to 40 (the recommended setting in the original implementation), and the chunk\_top-$k \in \{5,10,20,50\}$ .
For GraphRAG, we adopt the local retrieval mode, following the default configuration of the official implementation. 
DeepNote adopts the dense retrieval variant, using a vector-based retriever with top-$k \in \{5,10,20,50\}$ and max\_topk $\in \{15,30,60,150\}$. During reasoning, we set max\_step=3 and max\_fail\_step=2 to control iterative retrieval and failure tolerance.
RAT employs a Chroma-based dense retriever to retrieve from local corpus, also with top-$k \in \{5,10,20,50\}$. 
For ReAct, we use a vector retriever with the same top-$k$ range and set the reasoning iteration limit to 10 (max\_iteration=10) to control retrieval–generation alternation steps.

\subsubsection{Model information}
For all baselines that require text embeddings, we use text-embedding-3-small as the embedding model. 
For graph construction in LightRAG and GraphRAG, we employ GPT-4o-mini to generate entity and relation representations. 
Throughout the experiments, both reasoning and question-answering processes are conducted using GPT-4o-mini.

\begin{table*}[t]
\caption{Effect of retrieval Top-k on RAG baselines.}
\label{tab:rag-topk}
\centering
\small
\begin{tabular}{l c ccc ccc}
\toprule
\multirow{2}{*}{Baseline} & \multirow{2}{*}{Top-k} & \multicolumn{3}{c}{SpiderQA} & \multicolumn{3}{c}{AGGBench-Core} \\
\cmidrule(lr){3-5} \cmidrule(lr){6-8}
 & & NACE(Mean) & NACE(Med) & Recall(chunk) & NACE(Mean) & NACE(Med) & Recall(chunk) \\
\midrule
\multirow{4}{*}{Naive RAG} 
 & 5  & 0.666 & 0.800 & 0.181 & 0.463 & 0.500 & 0.042 \\
 & 10 & 0.670 & 0.800 & 0.364 & 0.598 & 0.429 & 0.066 \\
 & 20 & 0.657 & 0.800 & 0.521 & 1.039 & 0.444 & 0.093 \\
 & 50 & 0.672 & 0.800 & 0.403 & 1.855 & 0.500 & 0.072 \\
\midrule
\multirow{4}{*}{LightRAG} 
 & 5  & 0.700 & 0.750 & 0.768 & 0.930 & 0.500 & 0.033 \\
 & 10 & 0.672 & 0.692 & 0.848 & 0.868 & 0.500 & 0.051 \\
 & 20 & 0.708 & 0.750 & 0.865 & 0.902 & 0.500 & 0.096 \\
 & 50 & 0.740 & 0.750 & 0.867 & 0.888 & 0.500 & 0.147 \\
\midrule
\multirow{3}{*}{Deepnote} 
 & 5  & 0.425 & 0.162 & 0.827 & 0.621 & 0.667 & 0.015 \\
 & 10 & 0.643 & 0.111 & 0.898 & 0.785 & 0.667 & 0.037 \\
 & 20 & 0.776 & 0.098 & 0.933 & 0.963 & 0.615 & 0.049 \\
 & 50 & 0.429 & 0.137 & 0.965 & 0.920 & 0.556 & 0.094 \\
\bottomrule
\end{tabular}
\end{table*}

\subsection{Ablation Studies of DFA} \label{sec:ablation}
To understand the contribution of each component in the DFA pipeline, we conduct systematic ablation experiments.

\subsubsection{Ablation on Disambiguation Module} 

We conducted a detailed ablation study of the disambiguation module to assess its impact on the final query rewriting result. Collecting human clarification for every query would be prohibitively expensive, so we introduced an auxiliary model to simulate user feedback. Model~A handled the full disambiguation pipeline—including classification, clarification question generation, and final rewriting—while Model~B simulated user responses. Based on this simulated feedback, Model~A generated rewritten queries, and human annotators only needed to provide binary correctness judgments on the outputs, substantially reducing annotation costs. We compare two rewriting strategies: classification-guided rewriting and direct rewriting. As shown in Table~\ref{tab:rewriting}, classification-guided rewriting consistently outperforms direct rewriting, achieving accuracy gains of roughly 10\% across all models. These results confirm that explicitly identifying ambiguity types provides a stronger foundation for accurate query rewriting than attempting to rewrite in a single step. 

\begin{table}[t]
\centering
\caption{Disambiguation accuracy of two rewriting strategies. Classification-guided rewriting yields consistent improvements.}
\label{tab:rewriting}
\small
\setlength{\tabcolsep}{3pt}
\resizebox{\columnwidth}{!}{
\begin{tabular}{lcc}
\toprule
\textbf{Model A} & \textbf{Direct Rewriting (\%)} & \textbf{Classification-Guided (\%)} \\
\midrule
Gemini-2.5-Flash & 45.33 & 67.28 \\
GPT-4o mini      & 69.31 & 75.81 \\
DeepSeek-R1      & 69.92 & 80.49 \\
Qwen3 32B        & 63.62 & 77.44 \\
\bottomrule
\end{tabular}}
\end{table}


\subsubsection{Ablation on Filtering Module} We ablate the observation mechanism in the \emph{result evaluation} step of the filtering module. Specifically, we revert to a standard evaluation that inspects only the retained results and compare it with our method. On AGGBench and SpiderQA, the recall drops from 0.63 to 0.55 and from 0.89 to 0.85. These results confirm that incorporating observations of discarded evidence enables the model to better detect over-filtering and maintain completeness.

\begin{table}[t]
\centering
\caption{Ablation study results of the aggregation module.}
\label{tab:aggregation_ablation}
\resizebox{0.4\textwidth}{!}{
\begin{tabular}{lcc cc}
\toprule
\multirow{2}{*}{\textbf{Ablation}} & \multicolumn{2}{c}{\textbf{MusiQue}} & \multicolumn{2}{c}{\textbf{AGG-Core}} \\
\cmidrule(lr){2-3} \cmidrule(lr){4-5}
 & LC & CP & LC & CP \\
\midrule
w/o embed & 25 & 765 & 5 & 53 \\
w/o tfidf & 25 & 699 & 5 & 73 \\
w/o offline-cluster & 35 & 779 & 8 & 43 \\
w/o greedy-batching & 132 & 1874 & 51 & 247 \\
\textbf{Ours} & \textbf{25} & \textbf{569} & \textbf{5} & \textbf{41} \\
\bottomrule
\end{tabular}
}
\label{tab:ablation}
\end{table}
\subsubsection{Ablation on Aggregation Module}

We ablate each component of the aggregation module to examine its contribution. The results show an overall performance drop of about 20–70\% when individual components are removed. In particular, removing TF-IDF or embeddings weakens the feature extraction process, since TF-IDF measures whether chunks share similar keywords, while embeddings capture their semantic relatedness. Excluding offline clustering results in many fragmented clusters, and replacing semantic-aware greedy batching with traditional greedy prevents multiple clusters from being merged into the same batch, thereby increasing both LLM calls and candidate pairs.

\subsection{Effect of Increasing Retrieval Top-k}
For the first category of RAG-based approaches, a natural question arises: could simply increasing the retrieval window (Top-k) overcome their performance bottleneck? To examine this, we extended several representative baselines with larger retrieval sizes. While recall at the chunk level did improve, the overall aggregation accuracy exhibited little to no gain, and in some cases even degraded due to noise accumulation. These results demonstrate that merely enlarging Top-k is insufficient to resolve aggregation tasks, which directly motivates our design of the DFA framework.

\subsection{Can Larger Top-k Retrieval Solve Aggregation Problems?}
To verify whether enlarging the retrieval Top-k resolves the aggregation problem, we evaluated the performance of several RAG baselines under different Top-K. Results are summarized in Table~\ref{tab:rag-topk}. We observe that increasing Top-k raises chunk-level recall, sometimes by more than 0.9 compared to low-retrieval settings. However, this gain does not translate into better aggregation accuracy. On SpiderQA, naive RAG exhibits higher recall as top-k grows, yet both NACE(mean) and NACE(med) remain nearly unchanged, still far from DFA’s performance. LightRAG shows a similar trend: although recall on AGGBench increases from 0.03 to over 0.15, its NACE values fluctuate with larger top-k, reflecting unstable accuracy. For DeepNote, enlarging top-k yields smoother recall improvements, but aggregation accuracy also oscillates and eventually worsens at higher retrieval sizes, revealing the model’s sensitivity to noisy evidence.

Overall, results show that existing RAG methods cannot solve aggregation problems by enlarging Top-k. Even when evidence is retrieved, the inclusion of noise often undermines accuracy. This further demonstrates the necessity of our proposed DFA framework.